\documentclass[sigconf]{acmart}
\AtBeginDocument{%
  \providecommand\BibTeX{{%
    \normalfont B\kern-0.5em{\scshape i\kern-0.25em b}\kern-0.8em\TeX}}}

\setcopyright{acmcopyright}
\copyrightyear{2021}
\acmYear{2021}

\acmDOI{10.1145/1122445.1122456}

\acmConference[GLB 2021]{GLB 2021: Workshop on Graph Learning Benchmarks 2021}
\acmBooktitle{The Web Conference 2021}
\acmPrice{15.00}
\acmISBN{978-1-4503-XXXX-X/21/06}

\usepackage{multirow}
\usepackage{xurl}
\usepackage{hyperref}
\usepackage{url}

\begin{document}

\title{CandidateDrug4Cancer: An Open Molecular Graph Learning Benchmark on Drug Discovery for Cancer}

\author{Xianbin Ye}
\authornote{These authors contribute equally to this work.}
\author{Ziliang Li}
\authornotemark[1]
\affiliation{%
  \institution{Ping An Healthcare Technology}
  \city{Beijing}
  \country{China}
}

\author{Fei Ma}
\author{Zongbi Yi}
\affiliation{%
  \institution{Chinese Academy of Medical Sciences}
  \city{National Cancer Center, Beijing}
  \country{China}
}

\author{Pengyong Li}
\affiliation{%
  \institution{Xidian University
  \\ School of Computer Science and Technology}
  \city{Xian}
  \country{China}}

\author{Jun Wang}
\authornote{Corresponding author.}
\affiliation{%
  \institution{Ping An Healthcare Technology}
  \city{Beijing}
  \country{China
  \\  deeplearning.pku@qq.com}
}

\author{Peng Gao\\Yixuan Qiao}
\affiliation{%
  \institution{Ping An Healthcare Technology}
  \city{Beijing}
  \country{China}
}

\author{Guotong Xie}
\authornote{Corresponding author.}
\affiliation{%
  \institution{Ping An Healthcare Technology}
  \city{Beijing}
  \country{China
  \\xieguotong@pingan.com.cn}
}

\renewcommand{\shortauthors}{Ye and Li, et al.}


\begin{abstract}
Anti-cancer drug discoveries have been serendipitous, we sought to present the Open Molecular Graph Learning Benchmark, named CandidateDrug4Cancer, a challenging and realistic benchmark data- set to facilitate scalable, robust, and reproducible graph machine learning research for anti-cancer drug discovery. CandidateDrug4- Cancer dataset encompasses multiple most-mentioned 29 targets for cancer, covering 54869 cancer-related drug molecules which are ranged from pre-clinical, clinical and FDA-approved. Besides building the datasets, we also perform benchmark experiments with effective Drug Target Interaction (DTI) prediction baselines using descriptors and expressive graph neural networks. Experimental results suggest that CandidateDrug4Cancer presents significant challenges for learning molecular graphs and targets in practical application, indicating opportunities for future researches on developing candidate drugs for treating cancers. 
\end{abstract}




\begin{teaserfigure}
  \centering
  \includegraphics[width=0.76\linewidth]{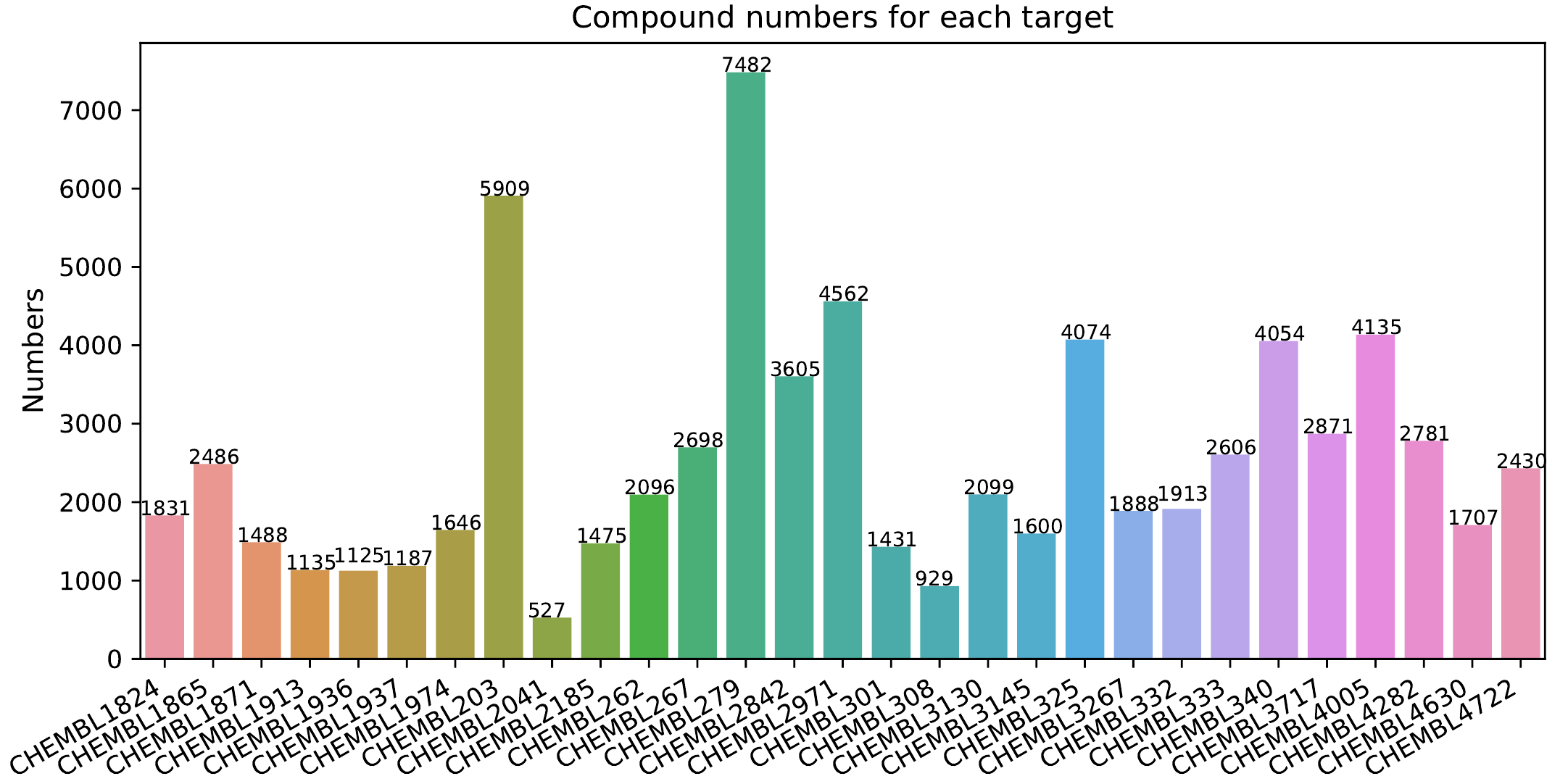}
  \vspace{-2ex}
  \caption{The CandidateDrug4Cancer benchmark encompasses the most-mentioned 29 targets for cancer, covering 54869 cancer-related drug molecules which is ranged from pre-clinical, clinical and FDA-approved.}
\end{teaserfigure}

\maketitle

\section{Introduction}
Cancer is one of the leading causes of mortality worldwide. Opportunities to help reduce the death rate from cancer through the discovery of new drugs are benefiting from the increasing advances in technology and enhanced knowledge of human neoplastic disease\cite{zhang2020overcoming,shaked2019pro}. However, drug discovery is a time-consuming, labor intensive, and expensive process, far slower than expected. The entire process from discovery to the regulatory approval of a new drug can take as much as 12 years and cost estimated at US 3 billion~\cite{dimasi2016innovation}. 
It has been usually hampered by experimental discovery of molecules and targets, and following with validation with in vitro experiments on cell lines and animals before moving to clinical testing~\cite{hill2012drug}.  Furthermore, stagnant success rate (~1:5000) is associated with each drug development stage.

Fortunately, many researchers have proposed various effective computer-aided drug discovery (CADD) methods~\cite{sliwoski2014computational} to decrease the costs and speed up projects~\cite{kapetanovic2008computer}. As an feasible assistant technique, CADD has made a significant contribution to drug discovery and has successfully developed dozens of drugs to market in recent decades\cite{yang2019concepts}.
Generally, CADD can be categorized into receptor-based methods and ligand-based methods upon the availability of target proteins and molecules. Receptor-based CADD relies on the target protein structure to calculate interaction for compounds, while ligand-based CADD exploits the information of compounds with diverse structures and known activity and inactivity, for construction of predictive, quantitative structure-activity relation (QSAR)\cite{cereto2015molecular}. 

\vspace{-2ex}

\begin{figure}[h]
  \centering
  \includegraphics[width=\linewidth]{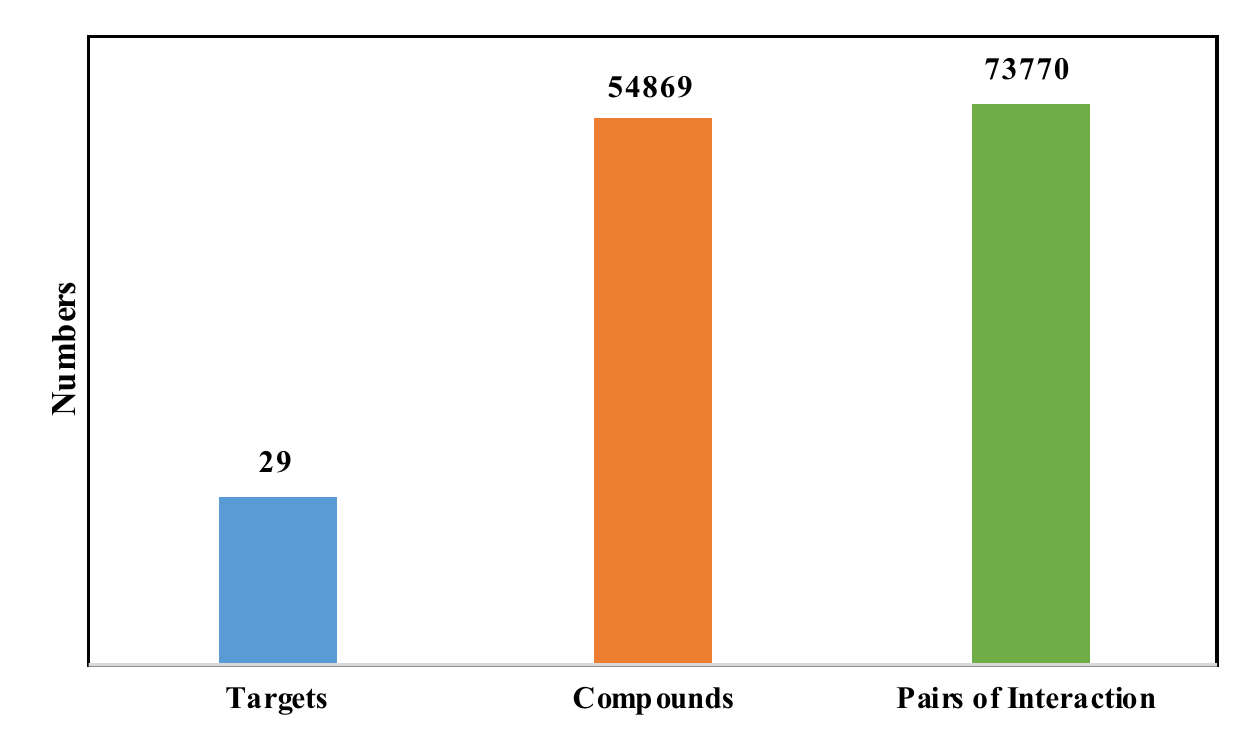}
  \vspace{-4ex}
  \caption{CandidateDrug4Cancer dataset encompasses multiple most-mentioned 29 targets for cancer, with 73770 drug-target pairs, covering a diverse range of 54869 related drug molecules which is ranged from pre-clinical, clinical and FDA-approved. }
  \Description{}
\end{figure}

\vspace{-2ex}

At present, the interdisciplinary studies between advanced Artificial Intelligence (AI) and drug discovery have received increasing attention due to the superior speed and performance. Many AI technologies have been successfully applied in a variety of tasks for CADD, such as receptor-based Drug Target Interaction (DTI) prediction~\cite{abbasi2020deep, d2020machine}.  Due to the large size of the chemical space, one of the fundamental challenges for these studies is how to learn expressive representations from molecular structures~\cite{yang2019analyzing}.
In the early years, molecular representations are based on hand-crafted features such as molecular descriptors or fingerprints~\cite{xue2000molecular}. 
In contrast, there has been a surge of interests in end-to-end molecular representation learned by graph neural networks. However, challenges for deep learning in molecular representation mainly arise from the scarcity of labeled data, resulting in 
models that lack practicability~\cite{hu2019strategies,rong2020grover}. In particular, so far there is limited published information on molecular graph learning for cancer to our knowledge. 

Altogether, these issues make it difficult producing favorable benchmark assembling proper information of cancer targets and potential molecules, for assessing and comparing the performance of different methods on drug discovery for cancer. To address these issues, we introduced an open molecular graph learning benchmark on drug discovery for cancer, named CandidateDrug4Cancer. Our main contributions can be summarized as follows:
\begin{itemize}
\item We introduce a novel graph learning task which has the potential to help understand the performance and limitations of graph representation facilitated models on candidate anti-cancer drug discovery problems. 

\item We provide CandidateDrug4Cancer benchmark dataset, which encompasses multiple most-mentioned 29 targets for cancer, covering 54869 cancer-related drug molecules which are ranged from pre-clinical, clinical and FDA-approved.

\item We conduct benchmark evaluations for drug target interaction with baseline encoders including powerful graph neural network for drug-like molecules.
\end{itemize}

\section{Related works}

\textbf{Drug Target Interaction} One of the initial steps of drug discovery is the identification of novel drug-like compounds that interact with the predefined target proteins. Various deep learning methods have been developed and achieved excellent performance for drug-target interaction (DTI) prediction~\cite{mousavian2014drug,chen2018machine,wen2017deep}. Generally, the deep learning algorithms for DTI prediction comprise of a compound encoder and a protein encoder. Recently, Tsubaki et al.~\cite{Tsubaki2019} proposed a new DTI framework by combining GNN for compounds and a CNN for proteins, which significantly outperformed existing methods.

\noindent\textbf{Molecular Graph Learning} Recently, among the promising deep learning architectures, graph neural network (GNN), such as message passing neural network (MPNN)~\cite{gilmer2017neural} has gradually emerged as a powerful candidate for modeling molecular data. Because a molecule is naturally a graph that consists of atoms (nodes) connected through chemical bonds (edges), it is ideally suited for GNN. Up to now, various GNN architectures have achieved great progress in drug discovery~\cite{Wu2018,li2021effective}. However, there are some limits that need to be addressed. Challenges for deep learning in molecular representation mainly arise from the scarcity of labeled data, as lab experiments are expensive and time-consuming. Thus, training datasets in drug discovery are usually limited in size, and GNNs tend to overfit them, resulting in learned representations that lack practicability~\cite{hu2019strategies,rong2020grover,lipairwise}.

\noindent\textbf{Anti-Cancer Candidate Drug Datasets} Unfortunately, the current number of drugs (FDA approved or at the experimental stage) is only around 10,000, the current knowledge about the drug–target space is limited, especially for severe life-threatening cancers, and novel approaches are required to widen our knowledge\cite{rifaioglu2020deepscreen}. However, most cancer drug discoveries have been serendipitous, the extent to which non-cancer drugs have potential as future cancer therapeutics is unknown\cite{corsello2020discovering}. Recent efforts have demonstrated the power of cancer cell line screening—testing either many compounds across a limited number of cell lines (for example, NCI-60 \cite{alley1988feasibility}, Genomics of Drug Sensitivity in Cancer (GDSC) \cite{garnett2012systematic}, the Cancer Target Discovery and Development (CTD2)\cite{basu2013interactive} and DepMap\cite{corsello2020discovering}. The ideal study would involve screening larger amount of drug candidates (most of which are non-oncology drugs) across most target proteins to capture the molecular diversity of human cancer.

\vspace{-2ex}

\begin{table}[htbp]
  \centering
  \caption{The most frequent target proteins of cancers collected from DepMap\cite{corsello2020discovering}, DrugBank\cite{wishart2008drugbank} and ChEMBL\cite{gaulton2012chembl}.}
    \begin{tabular}{ll}
    \toprule
    Target\_id & Target\_name \\
    \midrule
    chembl1824 & Receptor protein-tyrosine kinase erbB-2 \\
    chembl1865 & Histone deacetylase 6 \\
    chembl1871 & Androgen Receptor \\
    chembl1913 & Platelet-derived growth factor receptor beta \\
    chembl1936 & Stem cell growth factor receptor \\
    chembl1937 & Histone deacetylase 2 \\
    chembl1974 & Tyrosine-protein kinase receptor FLT3 \\
    chembl203 & Epidermal growth factor receptor erbB1 \\
    chembl2041 & Tyrosine-protein kinase receptor RET \\
    chembl2185 & Serine/threonine-protein kinase Aurora-B \\
    chembl262 & Glycogen synthase kinase-3 beta \\
    chembl267 & Glycogen synthase kinase-3 beta \\
    chembl279 & Vascular endothelial growth factor receptor 2 \\
    chembl2842 & Serine/threonine-protein kinase mTOR \\
    chembl2971 & Tyrosine-protein kinase JAK2 \\
    chembl301 & Cyclin-dependent kinase 2 \\
    chembl308 & Cyclin-dependent kinase 1 \\
    chembl3130 & PI3-kinase p110-delta subunit \\
    chembl3145 & PI3-kinase p110-beta subunit \\
    chembl325 & Histone deacetylase 1 \\
    chembl3267 & PI3-kinase p110-gamma subunit \\
    chembl332 & Matrix metalloproteinase-1 \\
    chembl333 & Matrix metalloproteinase-2 \\
    chembl340 & Cytochrome P450 3A4 \\
    chembl3717 & Hepatocyte growth factor receptor \\
    chembl4005 & PI3-kinase p110-alpha subunit \\
    chembl4282 & Serine/threonine-protein kinase AKT \\
    chembl4630 & Serine/threonine-protein kinase Chk1 \\
    chembl4722 & Serine/threonine-protein kinase Aurora-A \\
    \bottomrule
    \end{tabular}%
  \label{tab:addlabel}%
\end{table}%

\begin{figure*}[h]
  \centering
  \includegraphics[width=0.8\linewidth]{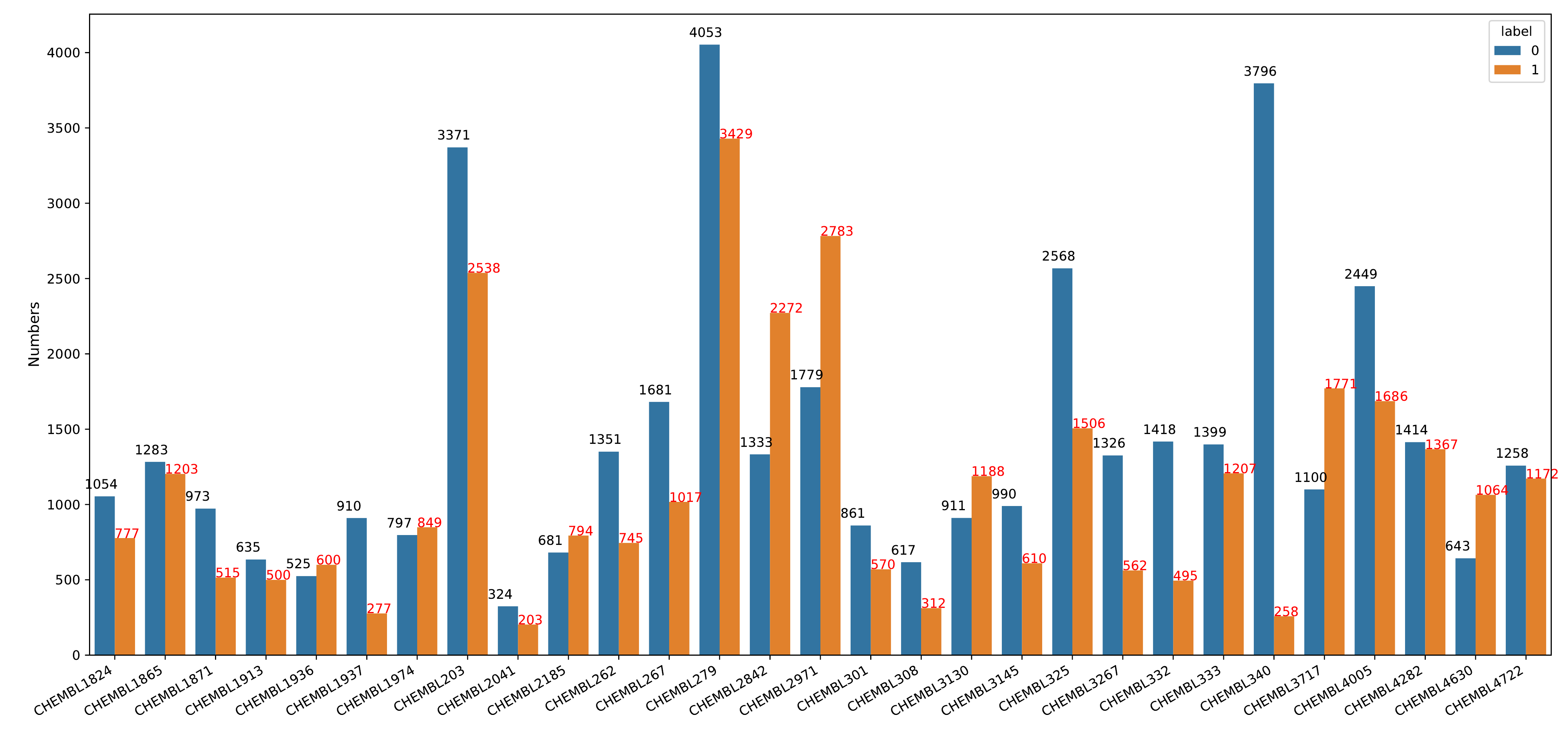}
  \caption{Label distribution of drug-target interaction pairs (boundary with pChEMBL at 7.0) across most-mentioned 29 targets for cancer in CandidateDrug4Cancer.}
  \Description{}
  \label{fig:label_count}%
\end{figure*}

\section{CandidateDrug4Cancer Datasets}

Inspired by DepMap \cite{corsello2020discovering} and DrugBank \cite{wishart2008drugbank}, we sought to create a public resource containing the candidate compounds for the most-used targets for cancers, as shown in Table \ref{tab:addlabel}. Firstly, the most commonly researched and cancer-related targets are collected based on DepMap and DrugBank, subsequently, the corresponding compounds and drug-target pairs are collected to establish CandidateDrug4Cancer from the ChEMBL database \cite{gaulton2012chembl} (https://www.ebi.ac.uk/chembl/g/, ChEMBL is a manually curated database of bioactive molecules with drug-like properties. It brings together chemical, bioactivity and genomic data to aid the translation of genomic information into effective new drugs). In addition to the published IC50 measurement, ChEMBL have added an additional field called pChEMBL to the activities table. This value represent comparable measures of concentrations to reach half-maximal response transformed to a negative logarithmic scale. pChEMBL is defined as:
\begin{equation}
    pChEMBL = -\lg (molar IC50(M))
\end{equation}

For example, an IC50 measurement of 10 uM would have a pChEMBL value of 5. However, in order to efficiently verify anti-cancer inhibitors and reduce the cost of a large number of sub-sequent experiments, we set decision boundary at pChEMBL is 7(approximately 100 nM) more strictly, defining pChEMBL larger than or equal to 7 as positive interaction \cite{lenselink2017beyond} (see Figure \ref{fig:label_count}).


\begin{table}[htbp]
  \centering
  \caption{The features used in molecular graph. These features are obtained by RDKit~\cite{rdkit2006}.}
  \resizebox{1\columnwidth}{!}{%
    \begin{tabular}{cccccccccc}
    \toprule
     Type & Name & Description   \\
          
    \midrule
    \multirow{9}[2]{*}{Node feature} 
          & Atom type & Atomic number (0-122)  \\
            & Formal charge & [unk,-5, -4, -3, -2, -1, 0, 1, 2, 3, 4, 5]  \\
                & Chirality type &[unk, unspecifie, tetrahedral-CW, tetrahedralL-CCW, other]  \\
              & Hybridization & [unk, sp, sp2, sp3, sp3d, sp3d2, unspecified]  \\
            & NumH &Number of connected hydrogens[unk,0, 1, 2, 3, 4, 5, 6, 7, 8] \\
              & Implicit valence &[unk, 0, 1, 2, 3, 4, 5, 6] \\
            
          & Degree &Number of covalent bonds [unk, 0, 1, 2, 3, 4, 5,6,7,8,9,10]   \\
          & Aromatic & Whether the atom is part of an aromatic system [0,1]\\
 
    \midrule
    \multirow{4}[2]{*}{Edge feature}
              & Bond direction & [None,endupright, enddownright] \\
          & Bond type & [Single, double, triple, aromatic] \\
          & Conjugation & Whether the bond is conjugated [0,1] \\
          & Ring &  Whether the bond is in Ring [0,1] \\
          & Stereo &  [StereoNone, StereoAny, StereoZ, StereoE] \\

    \bottomrule
    \end{tabular}%
    }
  \label{tab:featuret}%
\end{table}%


\vspace{-3ex}

\section{Experimental Setup}

\textbf{Baseline models}. Various deep learning methods have been developed and achieved promising performance for DTI prediction ~\cite{mousavian2014drug,chen2018machine,wen2017deep,Tsubaki2019}.Generally, the common DTI prediction models comprise of a compound encoder and a protein encoder. We have compared baselines using different combinations of compound and protein encoders including: MPNN-CNN, Daylight-AAC, Morgan-AAC and MolGNet-CNN (For compounds: Morgan fingerprints, Daylight-type fingerprints, MPNN on molecular graph. For proteins: Amino Acid Composition up to 3-mers, Convolutional Neural Network (CNN) on target sequence). In particular, we further adopt our pre-trained GNN named MolGNet~\cite{li2020learn}, to evaluate the effectiveness of pre-trained GNN on DTI prediction for caner targets. 

\noindent\textbf{Implementation details}. We conduct leave-one-target-out evaluation, the drug-target interaction pairs of the rest 28 targets are used for training, and the performances are calculated on each unseen target proteins. So as to evaluate whether models are capable of prediction on unseen targets. Due to hardware limitations and multiple baseline comparisons, we evaluate the baselines on 20\% of total datasets (the sub-datasets are also provided in our link). Each model is trained for 50 epochs. For graph learning, all drug molecules are pre-processed into hydrogen-depleted molecular graphs with nodes features, edge features, and adjacency matrix with RDKit~\cite{rdkit2006}. The detailed information about nodes features and edge features can be referred to Table \ref{tab:featuret}.


\vspace{-2ex}
\begin{figure*}[h]
  \centering
  \includegraphics[width=0.85\linewidth]{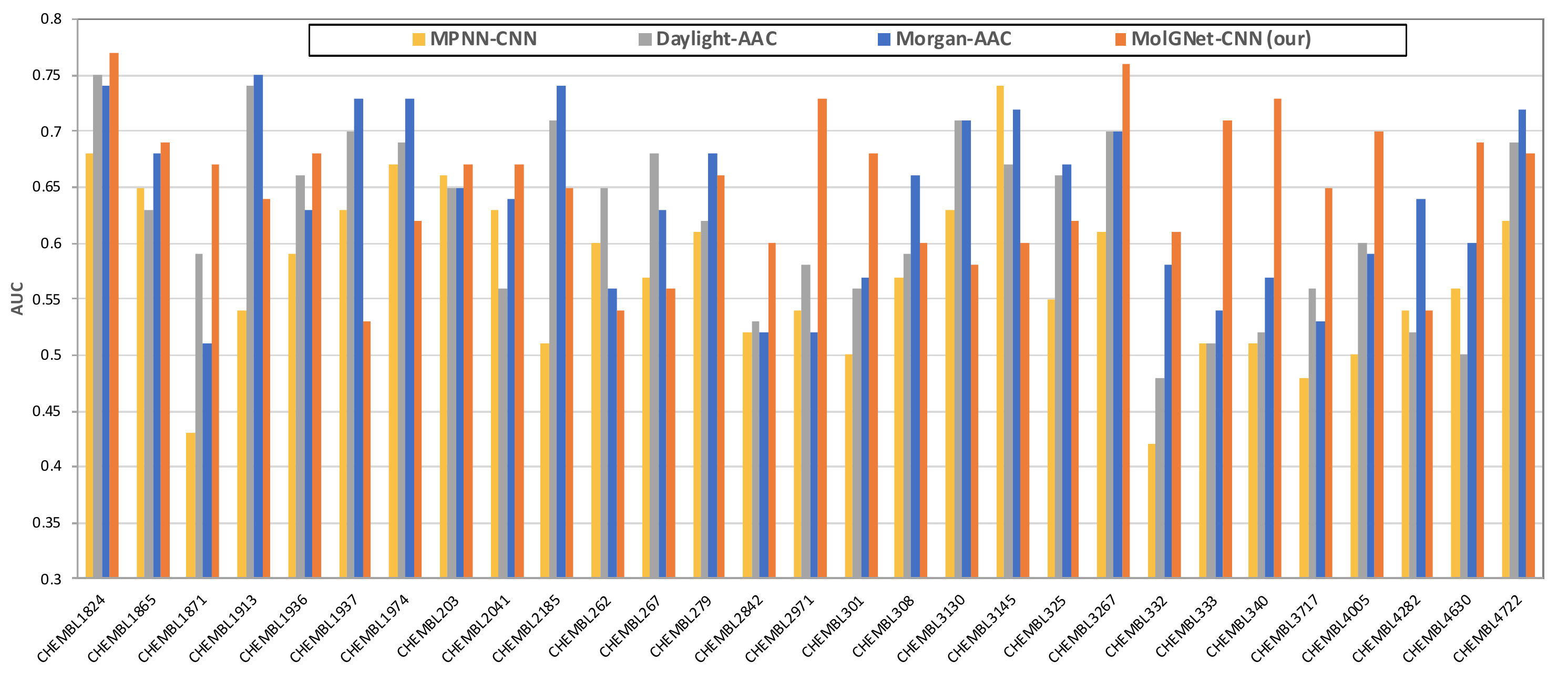}
  \caption{Baseline Comparison of the leave-one-out validation with DTI prediction AUC scores across 29 targets for cancer.}
  \Description{}
  \label{fig:result}
\end{figure*}

\section{Results and Conclusions}

As shown in Figure \ref{fig:result}, the average AUC scores of MPNN-CNN, Daylight-AAC, Morgan-AAC and MolGNet-CNN are 0.57, 0.62, 0.63, 0.65, respectively. The leave-one-out validation demonstrated that, the DTI model with expressive MolGNet encoder yielded superior performances on drug target interaction prediction , compared with other baselines. It re-confirmed graph learning as fundamental and powerful tools for modeling molecules. 

In summary, the CandidateDrug4Cancer is a starting point to develop new oncology therapeutics, and more rarely, for potential de novo drug design with powerful graph learning.
Despite the successes, there is still future improvement directions in the following aspects:
\textbf{Larger models and larger datasets}. It is interesting to employ even larger models and larger datasets for drug discovery on cancer. Larger models can potentially better handle more complicated learning.
\textbf{More suitable protein encoder}. It is desirable to explore more powerful protein encoders, so as to handle more complicated drug-target interaction.
\textbf{Weakly supervised learning}. It is also worthwhile to utilize semi-supervised learning and active learning techniques~\cite{wang2020semi} to further boost prediction performance. 
\textbf{Explainability of graph learning}. It is desirable to explore what useful insights or representations were learned in the graph embedding. Moreover, noting that DTI for cancer is still in the early stages but seen rapid growth/interest. Further analysis both theoretically and empirically is desired to better understand when/why/how graph learning can better work in drug discovery.

\vspace{-3ex}

\section{Online Resources}

The CandidateDrug4Cancer datasets and details are available at \protect\url{https://drive.google.com/file/d/1gXpGc5UhAYB9zVYnSl6E2MEaWfl5w98O/view?usp=sharing}.

\bibliographystyle{ACM-Reference-Format}
\bibliography{sample-base}


\end{document}